# How can NLP Help Revitalize Endangered Languages? A Case Study and Roadmap for the Cherokee Language

**Shiyue Zhang, Benjamin E. Frey, and Mohit Bansal**
UNC Chapel Hill
{shiyue, mbansal}@cs.unc.edu; benfrey@email.unc.edu

## Abstract

More than 43% of the languages spoken in the world are endangered, and language loss currently occurs at an accelerated rate because of globalization and neocolonialism. Saving and revitalizing endangered languages has become very important for maintaining the cultural diversity on our planet. In this work, we focus on discussing how NLP can help revitalize endangered languages. We first suggest three principles that may help NLP practitioners to foster mutual understanding and collaboration with language communities, and we discuss three ways in which NLP can potentially assist in language education. We then take Cherokee, a severely-endangered Native American language, as a case study. After reviewing the language's history, linguistic features, and existing resources, we (in collaboration with Cherokee community members) arrive at a few meaningful ways NLP practitioners can collaborate with community partners. We suggest two approaches to enrich the Cherokee language's resources with machine-in-the-loop processing, and discuss several NLP tools that people from the Cherokee community have shown interest in. We hope that our work serves not only to inform the NLP community about Cherokee, but also to provide inspiration for future work on endangered languages in general.[1]

## 1 Introduction

There are an estimated 6000 to 7000 spoken languages in the world, and at least 43% of them are endangered.[2] Throughout history, languages have naturally shifted and declined into dormancy. The current speed of language loss, however, is far beyond "natural". Some linguists estimate that between 50% and 90% of languages will be severely endangered or dead by the end of this century (Austin and Sallabank, 2011). This acceleration of language endangerment owes largely to cultural, political, and economic marginalization and the rise of global imperialism. Worldwide, indigenous people have suffered from colonization or conquest and given up their mother tongues in favor of another language. In order to achieve a higher social status, indigenous people have had to capitulate to colonizers' linguistic norms. Following Ladefoged (1992), we acknowledge that burdens such as raw material survival outweigh the more abstract concerns of maintaining a language. In other words, we cannot blame or fault indigenous people for giving up their languages in order to secure a better life under intense socioeconomic pressures. As linguists and NLP researchers, we have the responsibility to address these power imbalances and create a society where space exists for indigenous languages. Moreover, language loss is memory loss, identity loss, culture loss, and knowledge loss, and it even affects the health of indigenous people (Whalen et al., 2016).

Endangered languages are even more underrepresented in the NLP literature. Joshi et al. (2020) point out that more than 88% of the world languages spoken by around 1.2 billion people are *left behind*, i.e., they have been and are still ignored in the aspect of language technologies. Blasi et al. (2021) show that linguistic NLP tasks (e.g., morphology analysis) are more language inclusive than user-facing NLP tasks (e.g., machine translation). In this information age, NLP techniques are widely applied on the Internet. Much Internet content that we are exposed to daily is processed or even created by NLP techniques. Hence, the lack of NLP technology support for endangered languages reduces the degree to which users are exposed to them. Unfortunately, this exacerbates the problem of linguistic marginalization, as frequent language exposure is critical to language acquisi-tion. At worst, it can generate a downward spiral: since fewer speakers create content using these languages, the scarcity of resources will in turn hinder the development of NLP technologies. On the other hand, the majority of NLP research is biased towards high-resource languages, neglects diverse linguistic typologies (Joshi et al., 2020), and often relies on the availability of large-scale data. Including endangered languages can help diagnose NLP models' generalizability (Bender, 2011) and push towards universal and data-efficient approaches.

[1] Our code and data will be open-sourced at https://github.com/ZhangShiyue/RevitalizeCherokee.

[2] http://www.unesco.org/languages-atlas/en/statistics.html

In this work, we address three important steps on the roadmap of NLP for language revitalization: starting from "before NLP" to "NLP for language education" to "language-specific NLP research". Before diving into NLP research, we first suggest that NLP practitioners, who are often "outsiders" of indigenous communities, become aware of three important principles: *understand and respect first*, *decolonize research*, and *build a community*. We especially want to promote *building a community*. Since few people are speaking, learning, or studying an endangered language, the knowledge of each individual, the collected resources, and the developed models should be shared as widely and sustainably as possible. Hence, we need a community to support this (see Section 2).

Second, language revitalization is an attempt to reverse the decline of a language (Tsunoda, 2013). Fundamentally, this requires an increase in the number of active speakers to bring the language back to day-to-day use (Austin and Sallabank, 2011). Due to the lack of inter-generation transmission, language education in school or online is important. We introduce three approaches for applying NLP techniques in assisting language education (Section 3): *automated quiz generation*, *automated assessment*, and *community-based language learning*. The last approach connects to our previous point about building a community.

Next, we introduce the case study of Cherokee; an endangered[3] Native American language with only 2,000 fluent first-language speakers remaining. We first review its history (Section 4.1) to understand how social, political, and economic repression have harmed the Cherokee people and their language. Then, we discuss a few linguistic distinctions of Cherokee (Section 4.2), including polysynthesis, word order, etc., which can help us design linguistically informed NLP models. In Section 5, we review some existing high-quality Cherokee resources and propose two methods to enrich resources: *community-based resource collection* (which also relates to our previous point of building a community) and *automatic data mining*. Lastly, based on conversations with some Cherokee speakers/researchers, we dive deep into several NLP tools that seem advantageous for community members and may be able to create new usage domains for the language, and we point out the key challenges of their development (Section 6).

In summary, we propose suggestions to NLP practitioners, approaches of NLP-assisted language education, and directions for Cherokee language processing. We hope that our work can increase awareness of Cherokee and encourage more work on minority languages.

Last but not the least, the authors of this work come from both the Cherokee community (Benjamin E. Frey) and the NLP community (Shiyue Zhang and Mohit Bansal). Prof. Benjamin E. Frey is a proficient second-language Cherokee speaker and a citizen of the Eastern Band of Cherokee Indians. He has been teaching Cherokee and contributing to Cherokee revitalization for more than 10 years. He initiated our collaboration and continues bridging the gap between the Cherokee language and language technologies. In addition, we have been talking with some other Cherokee community members, including David Montgomery and Eva Marie Garroutte. Prof. Eva Marie Garroutte from Boston College said: "As a citizen of the Cherokee Nation, I am very concerned for the preservation of my tribe's endangered language and I am convinced that Dr. Frey's work represents the most promising project known to me for advancing this goal." Though by no means the views of this paper can represent the whole Cherokee community, our proposals are strongly initiated/motivated by Cherokee community members and grounded by NLP practitioners.

## 2 Before Diving into NLP Research

We suggest NLP practitioners, who are often "outsiders" of the indigenous communities, three general principles to follow before conducting NLP research on endangered indigenous languages.

**Understand and Respect First.** Meaningful advances in building speech and language technologies for under-resourced languages hinge upon be-ing able to understand those languages' speaker communities and their needs. Although the initial temptation among NLP researchers might be to dive in with questions about particular computational tools, that conversation cannot unfold until the speaker communities' more basic needs are met: *the need for respect, reciprocity, and understanding*. It may be tempting to say "this is outside the scope of our current research," yet these kinds of behaviors and assumptions are the very behaviors that led to the disenfranchisement of these groups. When we ignore someone's common humanity and assume that our need for control over the narrative and the situation is greater than their need to be seen and respected, we participate in the same marginalizing and dehumanizing behaviors that led to the problem we are purporting to address. Therefore, it is instrumental that we address the cultural practices and social norms of endangered language communities before assuming we know how to position ourselves, them, and our research within their communities.

[3] UNESCO has identified the dialect of Cherokee in Oklahoma is "definitely endangered", and the one in North Carolina is "severely endangered".

**Decolonize Research.** Decolonizing research is to place indigenous voices and needs in the center of the research process (Smith, 1999; Datta, 2018; Bird, 2020a). As NLP researchers, we are used to certain methodologies. When it comes to questions about endangered languages, it is tempting for us to formulate the new problems we encounter as what we are familiar with. However, we should always question ourselves: Is the formulation suitable for the language we conduct research on? Are the methodologies we familiar with the only true ways to solve the problems? Unquestioned focus on typical methodologies can make us treat languages as commodities and start to play a "number game" (e.g., the size of the data) and forget the real problem, language revitalization, we intend to solve in the first place (Dobrin et al., 2007). At every research step, it is critical to weigh the burden we put upon the speakers against the benefit that the research can bring back to their community. If the research outcome conveys no new knowledge, information, or benefit to the community, it is no different from "taking" indigenous knowledge that has occurred over the centuries. That is exactly why the word "research" is sometimes the direst (i.e., conjuring up bad memories) word in indigenous world's vocabulary (Smith, 1999). Finally, it is important to carefully deal with copyright and data governance; meanwhile, we advocate open-sourced and community-contributed works.

**Build a Community.** Fundamentally, we want to work together with people from the indigenous communities (Bird, 2020a, 2021). It is the most effective way to foster mutual understanding. We should communicate with the indigenous people and get to know their priorities. Common attitudes need to be fostered, common interests need to be found, and common goals need to be set up, before performing the research. These all lead to a community. We would imagine that there is an online community (a website) where native speakers can share their knowledge and language learners can find resources and learn the language together (see Section 3). People can share resources and participant in machine-in-the-loop resource collection projects (see Section 5). NLP researchers can evaluate and share their models in this community. Entertaining language learning or resource collection games can be launched. We hope the community can support wide and sustainable collaborations between indigenous speakers, language learners, and NLP practitioners. Compared to local communities of the speakers, this community will be greatly supported by technologies. A few NLP communities, e.g., MasakhaneNLP (focusing on African languages) and SIGEL (special interest group endangered languages), have been built. Differently, the community we promote here will support both NLP research and language learning. Lastly, compared to Telegram groups (we are in a few different Telegram groups with Cherokee community members), we want to build a more open community that everyone can have access to.

## 3 NLP-Assisted Language Education

Since little inter-generation language transmission is happening, language education is an essential requirement of language revitalization. Computer-assisted language learning has a long-standing history (Higgins, 1983) and two workshops, BEA[4] and NLP4CALL[5], are held for research on applying NLP for language education. Here, we discuss three ways in which NLP can potentially assist language education of endangered languages.

**Automated Quiz Generation.** The most direct way, in which NLP can help, is automatically generating quizzes for language learners. Practicing and producing the language in questions are critical to language acquisition (Gass and Mackey, 2013). Usually, language instructors manually design the quizzes, which is tedious and time-consuming; not to mention, there are not a lot of instructors for endangered languages. However, given the available text of endangered languages, NLP can easily and automatically generate cloze questions. It can also help find distracting wrong answers that happen in a similar context and thus form multi-choice questions (Hill and Simha, 2016; Susanti et al., 2018). To increase playfulness, language learning games, e.g., crossword puzzles and flashcards, can also be automatically generated (Rigutini et al., 2012; Xu and Ingason, 2021). Since these applications involve very basic language processing steps, NLP techniques can be reliably and easily applied.

[4] https://aclanthology.org/venues/bea/
[5] https://aclanthology.org/venues/nlp4call/

**Automated Assessment.** Another widely studied topic is NLP-supported automatic assessment. Though a lot of advanced assessments, e.g., grammar error correction (Bryant et al., 2019), essay grading (Chen et al., 2016), are difficult to be applied for endangered languages, we argue that some easier assessments are feasible. For example, automatic error analysis and template-based feedback can be provided for language learning quizzes. Another challenging but feasible assessment is to assess the readability or difficulty of language learning materials to provide suitable learning plans for learners of different levels. Using statistic and linguistic features, such as word frequency, morphology or syntactic complexity, etc., readability and difficulty can be automatically predicted (Schwarm and Ostendorf, 2005; Vajjala and Meurers, 2012). However, basic NLP tools, like POS tagger, dependency parser, morphology analyzer, need to be developed before these applications can be realized. The development of these tools requires small but highly-curated data (Blasi et al., 2021).

**Community-based Language Learning.** Free online language learning platforms that integrate automated quiz generation and assessment have been developed, e.g., Oahpa (Uibo et al., 2015). Taking one step further, we believe that a more effective approach of supporting endangered language education is to build an online and collaborative language learning platform, following the *human computation* technique (Von Ahn, 2008). When using technologies to assist in language revitalization, we often face a dilemma. On the one hand, due to the endangerment, there is not a lot of resources available and it is very expensive (in terms of time, effort, and cost) to collect resources from speakers. On the other hand, machines struggle to reach "useable" and "helpful" performances without a decent amount of training data. *Human computation* aims at combining human and computer to solve problems neither of them could solve alone (Von Ahn, 2008; Garcia, 2013). The most famous example is Wikipedia where Internet users contribute their knowledge together, and incredibly high-quality content has been created. Other successful cases are Duolingo and Tatoeba. Both are for language learners to translate web text and rate each other's translations. Then, the translated text can serve as learning materials and training data for NLP models. However, Tatoeba only has an English interface, and mixes languages on the same site, making it hard to find peer learners of under-resourced languages. Though Duolingo has language-specific sites, it supports 23 languages so far. Therefore, how to make use of collaborative language learning platforms for endangered languages is a big challenge. Nonetheless, we believe that it is a promising path to take for teaching endangered languages to the young generation in this information age.

## 4 The Cherokee Language

Starting from this section, we illustrate the situation of endangered languages through the example of Cherokee. We first review its history and linguistics. In the NLP area, we hardly get to know the languages and often let the model learn statistical patterns automatically from the data. However, it is critical to have basic knowledge of the language when contributing to its revitalization.

### 4.1 History of the Cherokee People and Their Language

**Tribal Sovereignty.** Before encountering Europeans, American Indians were already governing themselves. By drafting treaties with indigenous nations, the colonial powers implicitly recognized their sovereignty. Those treaties are still valid today, and tribal peoples are very much operating as sovereign nations, separate from the US (NCAI, 2020). There are three federally recognized nations of Cherokee people: Cherokee Nation of Ok-lahoma (CN), United Keetoowah Band of Cherokee Indians (UKB), and Eastern Band of Cherokee Indians (EBCI). Traditional Cherokee homeland covered parts of what are now eight US states.[6] EBCI is composed of those Cherokees who were able to remain in their homeland. CN is largely comprised of the descendants of those who were forcibly removed to Indian Territory along the infamous Trail of Tears in 1838 (Perdue and Green, 2007), while the UKB is composed largely of those whose ancestors chose to remove themselves west of the Mississippi. Although the three nations are politically independent, they all descend from the same Cherokee people, and maintain common interests, cultural elements, and language.

**The Language and its Dialects.** Cherokee is the only surviving member of the Southern Iroquoian language family, which have separated from the Northern Iroquoian languages about 4,000 years ago (Julian, 2010). James Mooney identified three main dialects of Cherokee: the Overhill dialect, the Underhill dialect (has died out), and the Middle, or Kituwah dialect. The Overhill dialect is primarily spoken in Oklahoma, and the Middle dialect is predominantly spoken in North Carolina today. Although according to UNESCO, both dialects are endangered, Cherokee is comparatively well-reported among American Indian languages. This is partially due to its writing system known as the 85-character Cherokee syllabary. It was invented in the early 1820s by Sequoyah (Britannica, 2021). The Cherokees have a newspaper written in their own language: the Cherokee Phoenix. The Phoenix, alongside the Cherokee New Testament, formed cornerstones of the Cherokee language in the 1800s on which many current language preservations and archiving projects rest.

**Language Endangerment.** Cherokee was robustly spoken until around the 1930s. The primary factor being responsible is the US government's "civilization" policy, which aimed to remove American Indians' cultural distinctions (Spring, 2016). Federal boarding schools were created on the model of military institutions by Richard H. Pratt under the philosophy of "kill the Indian, save the man" (Pratt, 2013). American Indian children were sent to residential schools to be educated in how to live in ways more similar to their white contemporaries. School overseers cut their hair, forced them to abandon their traditional dress, and punished them for speaking their traditional languages. Beyond the trauma, when they returned to communities, banks, post offices, factories, and grocery stores were all controlled non-locally. People working in them either no longer spoke Cherokee because they were not from Cherokee communities or because their employers were not Cherokee speakers. This transition contributed to the decline of the language in daily use, until the first generation grew up with only English as the language of the home around 1950s (Gulick, 1958; Frey, 2013). Recently, the larger project of language revitalization, of which this paper is a part, endeavors to return the language to regular day-to-day use in the Cherokee communities.

### 4.2 Cherokee Linguistics

**Polysynthetic.** Cherokee, like most American Indian languages, is polysynthetic. This means that words are primarily composed of a root whose meaning is modified by multiple prefixes and suffixes. The word ᎨᎦ, *gega*, can be divided up: *g-*, *-e-*, *-ga*. The *g-* prefix indicates that the subject of the verb is 1st person singular while the *-ga* suffix indicates that the action happens in the present tense and the aspect is progressive. The verb root *-e-* conveys the idea of motion. The simplest verb form in Cherokee will contain at minimum a root, a pronominal prefix, and a tense/aspect suffix. One oft-noted aspect of Cherokee grammar is its classificatory system, wherein verbs with direct objects must conjugate to indicate the physical shape of the direct object. The verb "I have," for instance, could appear in any of the following ways: *Agiha* (I have (solid)), *Agineha* (I have (liquid)), *Agwvya* (I have (long & rigid)), *Agina'a* (I have (flexible)), *Agikaha* (I have (animate)). Cherokee also has pre-pronominal prefixes that can specify the geographical location of particular events, such as *wi-* (translocative), which indicates that the action will happen at a distance away from the speaker, and *di-* (cislocative), which indicates the action will happen at a distance approaching the speaker.

**Word Order.** Word order in Cherokee is dependent on the larger pragmatic context in which the sentence appears, with new information or timeframes occurring before the verb and old or established information occurring post-verbally. Subject-object agreement is handled largely via the dual-argument pronominal prefixes. E.g., in “I see it,” ᏥᎪᏩᏘᎭ (*tsigowatiha*), the pronominal prefix *tsi-* indicates 1st person singular (“I”) acting on 3rd person singular (“it”). In ᎠᎩᎪᏩᏘᎭ (*agigowatiha*), we change *tsi-* to *agi-*, which means 3rd person singular acting on 1st person singular.

[6] North Carolina, South Carolina, Georgia, Kentucky, Tennessee, Alabama, Virginia, and West Virginia.

**Person & Number.** Although English has only two categories of number: *singular* and *plural*, Cherokee has a third, *dual* category. Therefore, a verb in Cherokee can be conjugated in first, second, or third person and specified for either singular, dual, or plural subjects. Dual and plural prefixes in the first person must then be further subdivided by clusivity, yielding 1st-person dual inclusive (you & I) or exclusive (she/he & I), 1st-person plural inclusive (all of us) or exclusive (they & I). The second person can inflect for dual (you two) or plural (you all). Cherokee does not have a third-person dual form, and speakers usually use the plural form when referring to two third persons.

**Verb-centric.** Cherokee is very verb-centric, and verbs comprise 75% of Cherokee (Feeling, 1975). Cherokee nouns are divided into root nouns (have no verbal inflection attached to them) and derived nouns (carry verbal morphology). Similarly, Cherokee adjectives can be distinguished from verbs in that their forms cannot carry the tense/aspect morphology typical of actual verbs. Thus, to say someone is skinny, ᎤᎴᏐᏓ (*ulesoda*) carries the pronominal prefix *u-*, indicating 3rd person singular, while ᎤᎴᏐᏓ ᎨᏒᎢ (*ulesoda gesv'i*) marks past tense by adding a separate copula (“to be”) that carries the tense/aspect suffix *-v'i*.

**Evidentiality.** Cherokee is also marked by a system of evidentiality (indicating whether one has firsthand knowledge of past events, or if one is reporting on hearsay). E.g., one might say ᎠᎦᏍᎬᎢ (*agasgv'i*), “it was raining (and I have firsthand knowledge of this)” vs. ᎠᎦᏍᎨᎢ (*agasge'i*), “it was raining (from what I understand).” Interestingly, this phenomenon applies regardless of the assumed truth of the statement in question.

**Phoneme.** Cherokee's phoneme inventory is, like other Iroquoian languages, almost completely bereft of bilabial sounds. It entirely lacks the *p* or *b* phonemes, along with *f*/*v*, *θ*/*ð*, and any *r* sound. It has six vowels: *a*, *e*, *i*, *o*, *u*, and *v*, and are generally pronounced with continental values, as in Spanish, except for *v*. Consonant inventory is small, at only 13, and most will be familiar to English speakers. The main exception is the voiceless alveolar fricative ɬ, likely more familiar to Icelandic speakers.

## 5 Cherokee Language Resources

The availability of language resources is not only important for language education but also determines the development of NLP technologies. Cherokee is categorized into “The Scraping-Bys” by Joshi et al. (2020), which means it has some amount of data but solid movements still need to be taken to increase the awareness of the language.

**Existing Resources Online.** It is not easy to locate a lot of Cherokee resources on the Internet, compared to other high-resource languages. Here, we point to a few places where high-quality Cherokee resources for language learning or NLP model training can be found: (1) Cherokee-English Dictionary[7] has online Cherokee-English dictionaries, a transliteration tool, a grammar guide, and a few Cherokee text or audio corpora; (2) Cherokee Nation website[8] contains Cherokee online classes, learning materials, fonts and keyboards, etc. (3) UNC Cherokee Program website[9] has UNC Cherokee class resources and pointers to external resources; (4) Cherokee Language Github group[10] gathers a lot of Cherokee text and audio data, as well as initial attempts for speech synthesis and some other NLP tools. (5) The Cherokee Phoenix[11] publishes all-Cherokee issues as well as some bilingual articles with Cherokee audios.[12] (6) We released around 17K Cherokee-English parallel data (Zhang et al., 2020).[13] In addition, Cherokee Wikipedia is available but its content is noisy. A Cherokee resource catalog can be built up in the future for easier locating resources.

**Community-based Resource Collection.** Besides existing resources, we suggest collaborative resource collection, which can be integrated with the community-based language learning platform we introduced in Section 3. A simple feature of this platform could be a dropbox where people who are willing to contribute their resources can drop in the files they have.[14] The back-end program can support any kind of data processing based on the contributor's request and permission. Then, the resources can be shared back with the community as language learning and model training resources. Second, for more complex data annotation tasks, like POS tagging, dependency parsing, we suggest setting up *game with a purpose* (GWAP) applications on this website. GWAP is introduced by Luis Von Ahn (Von Ahn, 2006; Von Ahn and Dabbish, 2008) who is also the founder of Duolingo. One famous example is his *ESP game* (Von Ahn and Dabbish, 2004) which formulates the *image recognition* task as a game. Following this idea, NLP practitioners can design diverse games on the platform to increase the fun and engagement of language learning and resource collection. In addition, this platform will focus more on what kind of materials the Cherokee community members consider important to preserve instead of what the NLP researchers find most valuable.

[7] `https://www.cherokeedictionary.net`

[8] `https://language.cherokee.org`

[9] `https://cherokee.web.unc.edu`

[10] `https://github.com/CherokeeLanguage`

[11] `https://www.cherokeephoenix.org`

[12] `https://tinyurl.com/4nf9txkf`

[13] `https://github.com/ZhangShiyue/ChrEn/tree/main/data/parallel_01172022`

**Automatic Resource Mining.** As NLP practitioners, we should try to make the most use of computers for collecting resources automatically. A lot of automatic *data mining* methods have been proposed to mine monolingual or bilingual text from the noisy web or Wikipedia (Guo et al., 2018; Artetxe and Schwenk, 2019; Schwenk et al., 2019; Wenzek et al., 2020; Schwenk et al., 2021; Arkhangelskiy, 2019). Though the mined text has many errors or noises, previous works demonstrate that neural NLP models are surprisingly good at using noisy data for training. However, some additional NLP components, like language identifier and multilingual embeddings, need to be developed to support the data mining. For instance, to mine Cherokee-English parallel text, we will need to map English and Cherokee sentences to the same representation space to compute their similarity. However, existing tools of getting multilingual sentence embeddings, like LASER,[15] do not support Cherokee, and Cherokee is not related to or sharing scripts with any supported languages. But, given the existing Cherokee-English parallel data (Zhang et al., 2020), we can re-train these tools and have Cherokee being supported. Note that these automatic text miners can start with both crawled web text and OCR-processed text (Section. 6.2).

[14] An example can be found at https://cherokee.web.unc.edu/submit-materials-to-database.

[15] https://github.com/facebookresearch/LASER

## 6 NLP Tools for Cherokee Language Processing

Based on our conversation with a few Cherokee speakers, they agree that some NLP tools are good to have and hold the potentials to be useful in Cherokee language revitalization. Thus, some initial attempts have been made by the Cherokee Language Github group and us (Zhang et al., 2020, 2021). Hence, we dive deep into several specific NLP tools for Cherokee language processing in this section. And for any NLP tool we develop, we want to evaluate it by the Cherokee speakers, and we suggest open-sourcing it for free usage. Connecting to our "build a community" proposal, we hope that NLP models can also be shared and used widely and sustainably in the community.

### 6.1 Machine Translation.

Ideally, a good machine translation (MT) system can automatically translate the big amount of English text to Cherokee; or it can assist human translators. Dr. David Montgomery, a citizen of Cherokee Nation and a Cherokee language learner, commented on MT: "It would be a great service to Cherokee language learners to have a translation tool as well as an ability to draft a translation of documents for first-language Cherokee speakers to edit as part of their translation tasks. If these tools can be made to work accurately, they would be transformative for the Cherokee language." Previously, we collected parallel text and developed an MT online translation demo between Cherokee and English (Zhang et al., 2020, 2021). However, our system can *translate fragments of the source sentence but make major mistakes*, which is far from being practically useful. The first challenge of MT development is the lack of data. Automatic data mining can help enrich MT training data (Section 5). But we still need high-quality and diverse evaluation data because existing evaluation sets (Zhang et al., 2020) are from limited domains (the majority is the Bible). Recently, Flores101, an MT evaluation benchmark covering 101 languages, has been created (Goyal et al., 2021). Though it has not yet covered Cherokee, we hope it can happen in the future.

The second challenge is processing and producing Cherokee text. Cherokee has rich morphology (see Section 4.2). One Cherokee word can be translated into one English sentence. Intuitively, we would think subword tokenization (Sennrich et al., 2016; Kudo, 2018) is helpful. However, previously, we (Zhang et al., 2020) showed that applying subword tokenization for English to Cherokee translation is harmful. We argue that it is because we processed Cherokee text in its syllabary rather than in transliterated Latin script, however, morphemes are easier to be learned from the latter. E.g., in ᏣᏆᏛᏍᏗᏒ, *tsaquadvsidisv* (when I was growing up), the prefix *ts-* marks relative clauses, but Ꮳ is *tsa*. We suspect that character-level generation (in Latin script) would work better for Cherokee. Additionally, Cherokee has flexible word order that is often determined by whether the information is new or old in relation to the larger discourse (Section 4.2). Thus, document-level translations are more reasonable than typical sentence-level translations.

| OCR tools | Original | | Screenshot | |
|---|---|---|---|---|
| | WER | CER | WER | CER |
| Tesseract | 0.355 | 0.230 | 0.151 | 0.063 |
| Google Vision | 0.533 | 0.199 | 0.468 | 0.074 |

Table 1: OCR performance of two OCR tools on our evaluation sets. WER: word error rate, CER: character error rates. For both WER and CER, lower is better.

### 6.2 Optical Character Recognition.

The majority of Cherokee text is in the format of manuscripts or books, so as many other endangered languages (Joshi et al., 2020; Bustamante et al., 2020). Though humans can read them, they are not machine-readable, which restricts the flexibility of their use, e.g., automatically creating language learning quizzes. Optical character recognition (OCR) (Smith, 2007) can help extract plain text from PDFs or images. Fortunately, existing OCR tools, like Tesseract-OCR[16] and Google Vision OCR API[17], support Cherokee and have decent accuracy. However, OCR accuracy is highly influenced by image quality. If the image has a noisy background or the text is surrounded by colorful pictures (which often happens in children books), the OCR accuracy will drop significantly.

To prove this, we create two evaluation sets from Cherokee books (including Cherokee New Testament, children books, Cherokee narratives): (1) *Original* has 20 images, and each image is one complete page from a book; (2) *Screenshot* is obtained by manually conducting screenshots and cutting out text from the 20 images, i.e., removing background noises. For each image in two sets, we manually annotate the corresponding text. Table 1 shows the results of Tesseract-OCR and Google Vision OCR API. Both OCR tools achieve significantly lower error rates on the *Screenshot* set than on the *Original* set, which demonstrates the importance of cleaning the images. Tesseract-OCR shows better performance than Google Vision OCR, especially it is better at detecting word boundaries. Although ways to improve image quality are available,[18] an easy-to-use tool need to be developed. OCR post-correction methods can also be applied (Rijhwani et al., 2020).

| | audio to phonetic text | audio to syllabic text |
|---|---|---|
| WER | 0.64 | 0.21 |

Table 2: The ASR results of finetuned XLSR-53 (Conneau et al., 2020) models. WER: word error rate.

### 6.3 Speech Recognition and Synthesis.

Automatic speech recognition (ASR) (Povey et al., 2011) can help language documentation, though indigenous community members may prefer unassisted transcription (Prud'hommeaux et al., 2021). Moreover, ASR holds the potential to automatically transcript audio data and thus enrich text corpus. A good amount of Cherokee audio data can be found from the "Cherokee Voices, Cherokee Sounds" radio, Cherokee Phoenix, and recorded meetings. ASR can automatically transcript these audios to produce valuable Cherokee text data. Recently, models that are first pre-trained on audio data and then finetuned on audio-text data have shown great advantages in performing ASR (Baevski et al., 2020). Especially, Conneau et al. (2020) pretrain and finetune a model on 53 languages and release XLSR-53 (supports ASR for 53 languages). It shows reasonable generalizability to unseen and low-resource languages. This sheds light on developing ASR for endangered languages.

Hence, we test its performance for Cherokee ASR. Using the audio-text data open-sourced[19] or shared privately by Michael Conrad, we build two ASR models: (1) audio to phonetic text, (2) audio to syllabic text. See more details in Appendix A.1. As shown in Table 2, we get surpris-ingly good performances, especially for the audio-to-syllabic-text model.[20] This is very promising, especially when knowing the fact that more self-training strategies can be applied, e.g., pretrain the speech encoder with Cherokee audio data, and more audio-text training data can be compiled. Text-to-speech synthesis (TTS) is more difficult to develop than ASR; nevertheless, following the pretrain-then-finetune paradigm, TTS models for extremely low-resource languages have been introduced (Xu et al., 2020).

[16] https://github.com/tesseract-ocr/

[17] https://cloud.google.com/vision/docs/ocr

[18] https://tinyurl.com/29xnewu9

[19] https://github.com/CherokeeLanguage/cherokee-audio-data

| | Precision | Recall | F1 |
|---|---|---|---|
| Unigram LM | 16.6 | 19.6 | 17.9 |
| BPE | 14.4 | 16.5 | 15.4 |
| Morfessor | 16.6 | 16.3 | 16.5 |

Table 3: The alignment between subwords and gold morphemes.

### 6.4 Tokenization and Morphology Parsing.

Tokenization is an essential pre-processing step of most NLP models, and it is related to morphology parsing. Subword tokenization has become *de facto* (Sennrich et al., 2016; Kudo, 2018). It segments a word into frequent subwords, and subwords are supposed to align with morphemes. Better alignment with morphemes can lead to better downstream performance (Bostrom and Durrett, 2020), while current subword tokenization methods struggle to perform well in morphologically rich languages (Amrhein and Sennrich, 2021).

Here, we evaluate how well subword tokenization can learn real morphemes for Cherokee. We train two subword tokenizers,[21] Unigram LM (Kudo, 2018) and BPE (Sennrich et al., 2016), and one morphology parser, Morfessor (Smit et al., 2014), on our previous MT training set (Zhang et al., 2020). Instead of using the original syllabic text, we transliterate text into Latin script to make it easier to learn morphemes. We collect gold (expert-labeled) morphemes of 372 Cherokee words from Cherokee Narratives (Feeling, 2018). Then, we use the pretrained tokenizers or parser to tokenize these 372 words and evaluate the alignment between subwords and gold morphemes. As shown in Table 3, subwords are poorly aligned with gold morphemes. Nonetheless, Unigram LM (Kudo, 2018) demonstrates better ability of inducing morphemes, which is consistent with the observation made by Bostrom and Durrett (2020). We think better representation methods need to be introduced for Cherokee, and the labeled data from Feeling (2018) can provide supervision.

[20]The same model finetuned on CommonVoice's Turkish data gets WER=0.35. https://tinyurl.com/62eykh9m

[21]We use SentencePiece (Kudo and Richardson, 2018).

### 6.5 POS-Tagging and Dependency Parsing.

More basic NLP tools like POS tagger and dependency parser are under-developed for Cherokee. These tools can not only support the development of other NLP tools but also be used to predict the readability of language learning materials (Section 3). Moreover, data for these tasks can serve as language learning materials for understanding Cherokee linguistics. Though unsupervised methods have been proposed (Stratos et al., 2016; Kim et al., 2019), usually small but high-quality labeled data, like Universal Dependencies (Nivre et al., 2016), is needed (Blasi et al., 2021). Therefore, data annotation by experts is required and community-based data collection strategies can be applied (Section 5). Moreover, the parallel English data and English tagger/parser can assist the annotation on the Cherokee side, which will also produce English-Cherokee word/phrase-level alignments as by-products. These alignments are valuable Cherokee language education resources, e.g., asking students when you have "structure X" in English, what is the corresponding "structure Y" in Cherokee?

## 7 Conclusion & Future Work

In this work, we discuss how NLP can help revitalize endangered languages. We first suggest general principles to NLP practitioners and propose ways of NLP-assisted language education. Especially, we promote building a (online) community that support collaborative language learning, resource collection, and knowledge sharing. Second, we conduct a case study for Cherokee (a severely-endangered Native American language). After reviewing Cherokee history and linguistics, we propose two methods of enriching Cherokee resources and discuss the developments of several NLP models that people from the Cherokee community are interested in. We hope our work can encourage future work to think and plan the path forward for other endangered languages. In the future, we hope to broaden our collaboration to even more Cherokee community members and build meaningful relationships with tribal governments, so that we can develop more useful applications through NLP techniques for supporting Cherokee revitalization.

## 8 Broader Impact Statement

The content of this paper is based on and inspired by our practice in Cherokee Language Revitalization. The conclusions and suggestions may or may not generalize to other endangered languages. For example, since Cherokee has its own syllabary and can be written down, we are interested in speech recognition for audio transcription. Even though some methods can directly translate audio to text of another language, we do not want to skip the transcription step. However, for some oral languages, they may want to prioritize translation over transcription to tackle the transcription bottleneck (Bird, 2020b). On the other hand, our position is influenced by Crystal (2014), who thinks using electronic technology is important for language revitalization. Therefore, a lot of our proposals, like "building an online community", may have an assumption that computers and the Internet have been or can be widely accepted and used in the indigenous community. However, it may not be true in every indigenous community.

## Acknowledgments

We thank the reviewers for their helpful comments. We thank Archiki Prasad and Zhiyuan Tang for providing guidance on developing ASR models. We thank Michael Conrad for providing Cherokee audios and transcriptions. We thank David Montgomery and Eva Marie Garroutte for providing their statements. We thank the Kituwah Preservation and Education Program (KPEP), the Eastern Band of Cherokee Indians, and the Cherokee Nation. This work was supported by NSF-CAREER Award 1846185, ONR Grant N00014-18-1-2871, NSF-AI Engage Institute DRL-2112635, and a Bloomberg Data Science Ph.D. Fellowship. The views contained in this article are those of the authors and not of the funding agency.

| Statistics | audio to phonetic text | | |
|---|---|---|---|
| | train | dev | test |
| Total Duration | 1.6h | 6.3m | 6.0m |
| Average Duration | 1.3s | 1.4s | 1.4s |

| Statistics | audio to syllabic text | | |
|---|---|---|---|
| | train | dev | test |
| Total Duration | 3h | 25.9m | 24.9m |
| Average Duration | 7.5s | 7.6s | 7.3s |

Table 4: The statistics of the data we use for developing the ASR models. s/m/h stands for second/minute/hour.

# A Appendix

## A.1 Data and Implementation Details of ASR

(1) audio to phonetic text: Given the audio, the model outputs text showing its pronunciation, e.g., *Sǔ:dáli* (means "six"). It follows Uchihara's Modified Community Orthography (Uchihara, 2013). (2) audio to syllabic text: Given the audio, the model outputs text showing the Cherokee syllabary, e.g., ᏑᏓᎵ (means "six").

We split our ASR data into training, development, and testing sets. Table 4 lists the statistics. It can be seen that we have more data for the audio to syllabic text model, which probably causes its good performance shown in Table 2.

We follow the ASR recipe provided by Huggingface's Transformers[22] (Wolf et al., 2020) to finetune the pretrained XLSR-53 model (Conneau et al., 2020). Specifically, we use learning rate=3e-4, epoch=15, mask_time_prob=0.01. We run each experiment for 3 times and report the average performance on the testing set in Table 2.

[22] https://github.com/huggingface/transformers/tree/master/examples/pytorch/speech-recognition